\documentclass[11pt,a4paper]{article}
\usepackage[hyperref]{emnlp-ijcnlp-2019}
\usepackage{times}
\usepackage{latexsym}

\usepackage{url}
\usepackage{microtype}
\usepackage{graphicx}
\usepackage{subfigure}

\usepackage{array}
\usepackage[font=small]{caption}
\usepackage{float}
\usepackage{dingbat}
\usepackage{multirow}
\usepackage{amsmath}
\usepackage{amssymb}
\usepackage{amsfonts}
\usepackage{makecell}
\usepackage{cleveref}
\usepackage{enumitem}
\usepackage{wrapfig}
\usepackage{natbib}
\usepackage{booktabs}
\usepackage{bbding}
\usepackage{tabularx}

\newcommand{\rar}{$\rightarrow$}

\aclfinalcopy 

\title{Countering Language Drift via Visual Grounding}

\author{Jason Lee$^\dagger$, Kyunghyun Cho$^{\dagger ^{\ddagger ^{\star}}}$, Douwe Kiela$^\ddagger$\\
$^\dagger$ New York University; $^\star$ CIFAR Azrieli Global Scholar; $^\ddagger$ Facebook AI Research\\
jasonlee@cs.nyu.edu, kyunghyun.cho@nyu.edu, dkiela@fb.com}

\begin{document}

\maketitle

\begin{abstract}
Emergent multi-agent communication protocols are very different from natural language and not easily interpretable by humans. We find that agents that were initially pretrained to produce natural language can also experience detrimental \emph{language drift}: when a non-linguistic reward is used in a goal-based task, e.g. some scalar success metric, the communication protocol may easily and radically diverge from natural language. We recast translation as a multi-agent communication game and examine auxiliary training constraints for their effectiveness in mitigating language drift. We show that a combination of syntactic (language model likelihood) and semantic (visual grounding) constraints gives the best communication performance, allowing pre-trained agents to retain English syntax while learning to accurately convey the intended meaning.
\end{abstract}

\section{Introduction}
A long-standing goal of artificial intelligence research is to develop agents that can cooperate with other agents, including humans, to solve tasks. As \citet{gauthier16paradigm} propose, one way to get closer to this goal is to develop agents that can flexibly use human language to coordinate with themselves and with humans. 

Recently, there has been a renewed interest in multi-agent communication~\citep{foerster2016learning,lazaridou2016multi}. While agents can be very effective in solving the tasks that they were trained on, their multi-agent communication protocols bear little resemblance to human languages. A major open question revolves around training multi-agent systems such that their communication protocols can be interpreted by humans.

One option is to pre-train in a supervised fashion with human language, but even then it is found that the protocols diverge quickly when the agents are fine-tuned on an external reward, as \citet{Lewis:17} showed on a negotiation task. Indeed, language drift is to be expected if we are optimizing for an external non-linguistic reward, such as a reward based on whether or not two agents successfully accomplish a negotiation.

\begin{table}[t]
    \begin{tabular}{lp{3.5cm}}
    \toprule
    Intended message: & \emph{2 elephants and 1 lion}\\
    \midrule
    No constraints & \emph{floopy globber}\\
    Syntactic & \emph{democracy is a political system}\\
    Syntactic+Semantic & \emph{a pair of elephants and a large feline}\\
    \bottomrule
    \end{tabular}
    \caption{Examples of valid messages under different constraints. Without constraints, agents may invent an arbitrary protocol to communicate the intended message. Syntactic constraints enforce ``Englishness'' but not semantic correspondence. Semantic constraints, e.g. with visual grounding, can enforce the communication channel to retain the meaning of the message.}
    \label{tab:illustration}
\end{table}

Language drift might be avoided by imposing a ``naturalness'' constraint, e.g. by factoring language model likelihood into the reward function. However, such a constraint only acts on the \emph{syntax} of the generated language, ignoring its \emph{semantics}. See Table~\ref{tab:illustration} for an illustration of different constraints. As has been advocated by multi-modal semantics \cite{Baroni:16,Kiela:17thesis}, we investigate if appropriate semantic constraints can be imposed on the generated language through (in this case visually) \emph{grounding} its meaning in a different modality.

In order to carefully study this problem, we require a task where drift can be accurately measured. Inspired by~\newcite{Lee:18}, we use a multi-modal machine translation (MMT) dataset \cite[Multi30k;][]{Elliott:16} to construct a new communication game: Two machine translation agents---i.e., encoder-decoder models with attention---are tasked with successfully translating source language sequences to the target language using a third pivot language as an intermediary. The first agent's decoder output is fed into the second agent's encoder as input. We employ policy gradient methods to train the first agent with the target language log-likelihood as reward. Thus, we effectively fine-tune two pre-trained machine translation agents via a pivot language, facilitating the study of its drift.

Contrary to alternative two-agent communication tasks such as navigation, game-playing or dialogue---which either don't have clearly defined metrics or easily available natural language data---this pivot-based translation allows us to check exactly whether the communicated sequence corresponds to the intended meaning, as well as to the gold standard sequence. In addition, every single utterance has very clear and well-known metrics such as BLEU and log-likelihood, allowing us to measure performance at every single step.

In what follows, we show that language drift happens, and quite dramatically so, when fine-tuning using policy gradients. Next, we investigate imposing syntactic conformity (i.e., ``Englishness'') via language model constraints, and show that this does somewhat mitigate drift, but does not lead to semantic correspondence. We then show that additionally imposing semantic constraints via (visual) grounding leads to the best retention of original syntax and intended semantics, and minimizes drift while improving performance. We conduct a token frequency analysis, which corroborates our hypothesis, and show that grounding causes the model to better preserve the token frequency distribution of the pivot language (English), while fine-tuning with language model constraints alone leads to a frequency distribution different from the original natural language.

The ability to keep drift in check opens up exciting possibilities for natural language processing research: we could maximize reward while \emph{retaining} the ``Englishness'' of the decoder, with obvious benefits for interpretability and interaction with humans. One general use case would be fine-tuning a language model pre-trained on large amounts of data for a given generation task with limited data, which is especially interesting given the recent interest in pre-trained language models \cite{Radford:2019gpt}. 
For instance when training chit-chat dialogue agents, we often want to optimize for some very high-level reward, such as engagingness or consistency, with hardly enough data to learn simple English grammar. The ability to fine-tune a pre-trained independent ``language module'', without drift, is an exciting prospect. With this work, we aim to take a step in that direction, and show that semantic constraints in the form of grounding play an important role.

\section{Prior Work}

Our work is inspired by recent work in protocols or languages that emerge from multi-agent interaction~\citep{Lazaridou:17,Lee:18,Andreas:17,Evtimova:18,Kottur:17,Havrylov:17,Mordatch:17}. Work on the emergence of language in multi-agent settings goes back a long way \citep{Steels:97,Nowak:99,Kirby:01,Briscoe:02,Skyrms:10}. In our case, we are specifically interested in \emph{tabula inscripta} agents that are already pre-trained to generate natural language, and we are primarily concerned with keeping their generated language as natural as possible during further training.

Reinforcement Learning has been applied to fine-tuning models for various natural language generation tasks, including summarization~\citep{Ranzato:15,Paulus:17}, information retrieval~\citep{Nogueira:17}, MT~\citep{Gu:17,Bahdanau:16} and dialogue~\citep{Li:17}. Our work can be viewed as fine-tuning MT systems using an intermediary pivot language. In MT, there is a long line of work of pivot-based approaches, most notably \citet{Muraki:86} and more recently with neural approaches \citep{Wang:17,Cheng:17,Chen:18}. There has also been work on using visual pivots directly \citep{Hitschler:16,Nakayama:17,Lee:18}. Grounded language learning in general has been shown to give significant practical improvements in various natural language understanding tasks~\citep{Gella:17,Elliott:17,Chrupala:15,Kiela:17groundsent,Kadar:18}. 

\section{Task and Models}

\subsection{Communication Task}

\begin{figure}[t]
\includegraphics[width=7.5cm]{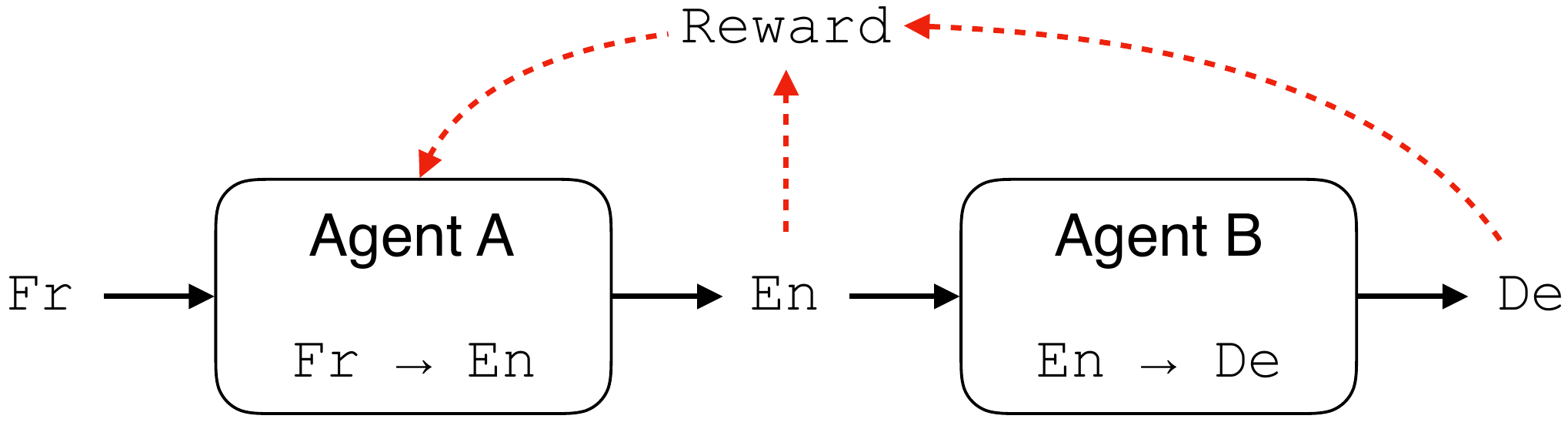}
\centering
\caption{Diagram of our communication game.}
\label{fig:diagram}
\end{figure}

We recast pivot-based translation as a communication game involving two MT agents: Fr\rar En and En\rar De. See Figure~\ref{fig:diagram}. Our dataset consists of $N$ triples of aligned sentences $\{\text{Fr}_k, \text{En}_k, \text{De}_k\}_{k=1}^N$. Note that $\text{En}_k$ is only used for evaluation and is not required for training. We first feed the French sentence $\text{Fr}_k$ to Agent A, which generates an English message $\overline{\text{En}_k}$ as output. Agent B is then trained to maximize the log-likelihood of the ground truth German sentence given the English message, i.e. $\log{p(\text{De}_k | \overline{\text{En}_k})}$. Agent A is trained using REINFORCE~\citep{Williams:92} with reward $R = \log{p_B(\text{De}_k | \overline{\text{En}_k})}$.\footnote{We use subscript $B$ to denote that the probability is computed with Agent B.} This encourages Agent A to develop helpful communication policies for Agent B, and allows Agent B to adapt to Agent A's new policies. In other words: communication via the pivot language (English) is a success if we are able to translate the intended source sequence (French) into the desired target sequence (German).

Both agents are pre-trained on their respective tasks before communication, which means that the intermediate language starts off as English in the early stages of the communication game, where the goal is to translate French to German. This work examines what happens to the intermediate language as we fine-tune the system jointly for the given goal: will the agents keep communicating in English, or diverge? And if so, what can we do to prevent that from happening?

This particular task and setup directly addresses the problem of language drift, as the availability of ground truth references and well-understood metrics (e.g. BLEU) allows us to exactly measure the degree of language drift over time. The Fr\rar En\rar De BLEU score informs communication success, while (the relative change in) the Fr\rar En BLEU score captures the degree of language drift.

\subsection{Constraints via Auxiliary Tasks}

The action space of Agent A is $|V|^{L}$, where $|V|$ is the size of the vocabulary (approximately 20k) and $L$ is the sequence length. We explore the two aforementioned constraints: a syntactic constraint via language modeling (LM) and a semantic constraint via grounding (G).

\paragraph{Language Model (LM)} Given a language model pre-trained on a standard English corpus, the (sentence-level) log-likelihood of the English message informs its general ``Englishness''. We incorporate this into the reward for Agent A, so that it learns to send messages that are plausible English.\footnote{We also experimented with a dense LM reward on the word-level, but found this to lead to worse performance. We hypothesize that the model might be focusing too much on the dense LM reward, ignoring the sparse reward for the communication task and leading to poor performance. We did not use BLEU as it is a corpus-level metric.} Reward for Agent A is: $$R^\text{LM}_k = \log{p_{B}(\text{De}_k | \overline{\text{En}_k})} + \beta_{LM}\,\log{p_{LM}(\overline{\text{En}_k})}.$$ 

\paragraph{Grounding Model (G)} Let us assume we have access to a set of images $\{\text{Img}_k\}$ associated with each triple $\{\text{Fr}_k, \text{En}_k, \text{De}_k\}$. Given a pre-trained image-caption retrieval model, such as VSE++~\citep{Faghri:18}, the log-likelihood of the image given the English message (and vice versa) captures how much the English message is grounded in the original semantic content~\citep{Kiela:17groundsent}. We incorporate the ranking loss into Agent A's reward. $$R^\text{G}_k = \log{p_{B}(\text{De}_k | \overline{\text{En}_k})} + \beta_{G}\,\log{p_{G}(\text{Img}_k|\overline{\text{En}_k})}.$$ 
$\beta_{LM}$ and $\beta_{G}$ are hyperparameters.

\subsection{Training Objective}

Let us denote the $t$-th token in the $k$-th English training example with $\text{En}_{k}^{t}$, the actual reward and the state-dependent baseline in the $k$-th training example as $R_k$ and $\overline{R_{k}^t}$.


\paragraph{Policy Gradient Training} At decoding timestep $t$, Agent A takes an action (outputs token $\overline{\text{En}_{k}^t}$) given an environment (previous hidden states and previous token $\overline{\text{En}^{t-1}_k}$). It receives reward $R_k$ at the end of the sequence, from which we subtract a state-dependent baseline $\overline{R^t_k}$ to reduce variance. Therefore, we maximize $(R_k - \overline{R^t_k})\log{p( \overline{\text{En}^t_k} | \overline{\text{En}^{<t}_k}, \text{Fr}_k )}$. In addition, we employ entropy regularization on Agent A's decoder to encourage exploration. Hence, Agent A's overall objective function is given as: 
\begin{equation*}
\begin{split}
\mathbb{L}_A &= \sum_{k=1}^N \bigg\{ \sum_{t=1}^{T_k} \Big\{ 
\alpha_{\text{entr}} \, H\big(p( \overline{\text{En}^t_k} | \overline{\text{En}^{<t}_k}, \text{Fr}_k )\big) \\
&- \alpha_{\text{b}} \, \text{MSE}(R_k, \overline{R^t_k}) \\
&+ \alpha_\text{pg}\,(R_k - \overline{R_k^t})\log{p( \overline{\text{En}_k ^t} | \overline{\text{En}^{<t}_k}, \text{Fr}_k )} \Big\} \bigg\}, 
\end{split}
\end{equation*}
 where $H$ and $\text{MSE}$ denote entropy and mean squared error losses. $T_k$ is the maximum decoding timestep in the $k$-th training example.

\paragraph{Cross Entropy Training} Agent B is trained using standard cross entropy loss, i.e. $$\mathbb{L}_B = \sum_{k=1}^N \Big\{ \sum_{t=1}^{T_k} \log{p(\text{De}^t_k | \text{De}^{<t}_k, \overline{\text{En}_k})} \Big\}.$$

We jointly train both agents by maximizing $\mathbb{L} = \mathbb{L}_A + \mathbb{L}_B$.

\section{Experimental Settings}

In this section we provide the details of our experimental setup: a Fr\rar X\rar De translation task where the intermediate language X is initialized as English, and subsequently fine-tuned with policy gradient methods. The model is trained either with no constraints (\textbf{PG}), syntactic constraints via language modeling (\textbf{PG+LM}) or both syntactic and semantic constraints via language modeling and grounding (\textbf{PG+LM+G}).

\paragraph{Datasets} Agents A and B are initially pre-trained on IWSLT Fr\rar En and En\rar De, respectively \cite{Cettolo:2012iwslt}. Fine-tuning is performed on Multi30k Task 1~\citep{Elliott:16}. That is, importantly, there is no overlap in the pre-training data and the fine-tuning data. Multi30k Task 1 consists of 30k images and one caption per image in English, French, German and Czech (of which we only use the first three). To ensure our findings are robust, we compare four different language models, trained on WikiText103, MS COCO, Flickr30k and all of the above.

The grounding model is trained on Flickr30k \cite{Young:2014flickr30k}. Following \citet{Faghri:18}, we randomly crop training images for data augmentation. We use 2048-dimensional features from a pretrained and fixed ResNet-152~\citep{He:16}.

\paragraph{Preprocessing} The same tokenization and vocabulary are used across different tasks and datasets. We lowercase and tokenize our corpora with Moses~\citep{Koehn:07} and use subword tokenization with Byte Pair Encoding (BPE)~\citep{Sennrich:16} with 10k merge operations. This allows us to use the same vocabulary across different models seamlessly (translation, language model, image-caption ranker model).

%
%

\paragraph{Controling the English message length} When fine-tuning the agents, we observe that the length of English messages becomes excessively long. As Agent A has no explicit incentive to output the end-of-sentence (EOS) symbol, it tends to keep transmitting the same token repeatedly. While redundancy might be beneficial for communication, excessively long messages obscure evaluation of the communication protocol. For instance, BLEU score quickly deteriorates as the message length becomes longer, as it is a precision metric. When the message length is fixed, a drop in BLEU score will by necessity mean that the intermediate language has drifted away more. For this reason, we constrain the length of English messages to be no longer than the length of their French source sentence, or shorter if the model outputs the EOS symbol early.  Recall that Agent B is supervised to predict the EOS symbol at the right position, so does not suffer from this issue. 

\paragraph{Model Architecture and Pretraining} Our MT agents are standard sequence-to-sequence models with attention~\citep{Bahdanau:15} with a unidirectional, 1-layer GRU~\citep{Cho:14gru} with 256 hidden units and 256-dimensional embeddings. During initial pre-training on IWSLT, we early-stop on the validation BLEU score (tst2013). The best checkpoints give 34.05 BLEU and 21.94 BLEU on IWSLT Fr\rar En and En\rar De development sets with greedy decoding. For our policy gradient value function, we use a 2-layer MLP with ReLU activations.

The language model is a 1-layer recurrent language model with 512 LSTM hidden units. The image-caption retrieval model is a recently proposed VSE++ model~\citep{Faghri:18}, with a unidirectional 1-layer GRU with 512 hidden units and a single fully connected layer from 2048-dimensional ResNet features to 512-dimensional GRU hidden states. 

\paragraph{Training Details} When fine-tuning our agents, we perform learning rate annealing and early stopping based on Fr\rar En\rar De BLEU (communication performance) on the Multi30k development set. We use Adam~\citep{Kingma:14} with an initial learning rate of 0.001 and dropout~\citep{Srivastava:14} rate of 0.1. We grid search over the learning rate schedule and the reward coefficients ($\alpha_\text{pg}, \alpha_\text{entr}, \alpha_\text{b}$) for agent A and ($\beta_{LM}, \beta_{G}$) for agent B, respectively (see previous section). For our joint systems with policy gradient fine-tuning, we run every model three times with different random seeds and report averaged results.

\begin{table*}[t]
    \small
    \begin{center}
    \begin{tabular}{ l l c l l } \toprule
                    & \multicolumn{1}{c}{LM} & \multicolumn{1}{c}{Ranker} & \multicolumn{1}{l}{A: Fr\rar En} & \multicolumn{1}{l}{A\&B: Fr\rar En\rar De} \\ \midrule
        \multicolumn{2}{l}{Pretrained} &  & 27.18 & 16.30 \\
        \multicolumn{2}{l}{Ensembling} &  &  & 16.95 \\
        \multicolumn{2}{l}{Agent A fixed } & & 27.18 & 22.37 \\ \midrule
        \multirow{1}{*}{PG} & No LM & & 12.38 (0.67) & 24.51 (1.48) \\ \midrule
        \multirow{1}{*}{PG+G} & No LM & \CheckmarkBold  & 14.20 (1.58) & 26.23 (1.08) \\ \midrule
        \multirow{4}{*}{PG+LM} & WikiText103 &  & 21.63 (1.25) & 26.88 (0.12) \\
                                & MS COCO & & 25.05 (1.40) & 27.66 (0.34) \\
                                & Flickr30k &  & 24.85 (1.14) & 27.60 (0.27) \\ 
                                & All &  & 23.60 (1.05) & 27.67 (0.39) \\ \midrule
        \multirow{4}{*}{PG+LM+G} & WikiText103 & \CheckmarkBold  & 23.65 (1.91) & 27.87 (0.15) \\
                                & MS COCO & \CheckmarkBold  & 26.24 (0.28) & 27.86 (0.24) \\
                                & Flickr30k & \CheckmarkBold  & 25.99 (1.62) & 27.82 (0.41) \\ 
                                & All & \CheckmarkBold  & 24.75 (0.40) & 28.08 (0.73) \\ \midrule
        Fr\rar De   & & & & 30.73 \\
    \bottomrule
    \end{tabular}
    \caption{Results in BLEU score on Multi30k Task 1. For our models using policy gradient fine-tuning, we report results averaged over three runs and provide standard deviations in brackets. \textbf{PG} (no constraint): trained with vanilla policy gradient fine-tuning. \textbf{PG+G} (semantic): trained with grounding only. \textbf{PG+LM} (syntactic): trained with ``Englishness'' constraint. For MS COCO and Flickr30k, the LM was trained directly on image captions. \textbf{PG+LM+G} (syntactic+semantic): trained with grounding loss as well as the LM loss. Fr\rar En: degree of intermediate language drift in agent A; lower indicates more drift. Fr\rar En\rar De: overall A\&B communication performance; higher is better. For LM=All the LM was trained on all three LM datasets combined.}
    \label{tab:quan_results}
    
    \vspace{0.2cm}
    \begin{tabular}{l c c c c c} \toprule
            &    & \multicolumn{4}{c}{PG+LM} \\
            & No LM &  WikiText103 & MS COCO & Flickr30k & All \\
    +G & \CheckmarkBold & \CheckmarkBold & \CheckmarkBold & \CheckmarkBold & \CheckmarkBold \\
    \bottomrule
    \end{tabular}
    \caption{Using the bootstrapped Wilcoxon signed-rank test~\citep{Wilcoxon:45}, Fr\rar En results of PG+LM+G are found to be significantly different from its baselines in all cases considered (on all LM datasets) within the threshold of $p=0.02$.}
    \label{tab:wilcoxon}
    \end{center}
\end{table*}

\paragraph{Baseline and Upper Bound} Our main quantitative experiment has three baselines:
    \begin{itemize}
        \item Pretrained models : models pretrained on IWSLT are used without finetuning.
        \item Ensembling : Given $\text{Fr}$, we let Agent A generate $K$ En hypotheses with beam search. Then, we let Agent B generate the translation $\overline{\text{De}}$ using an ensemble of $K$ source sentences~\citep{firat2016zero,zoph2016multi}.
        \item Agent A fixed : We fix Agent A (Fr\rar En) and only fine-tune Agent B using $\mathbb{L}_B$. This shows the communication performance achievable when Agent A cannot drift.
    \end{itemize}
\begin{figure*}[!t]
\centering
\includegraphics[width=.28\textwidth]{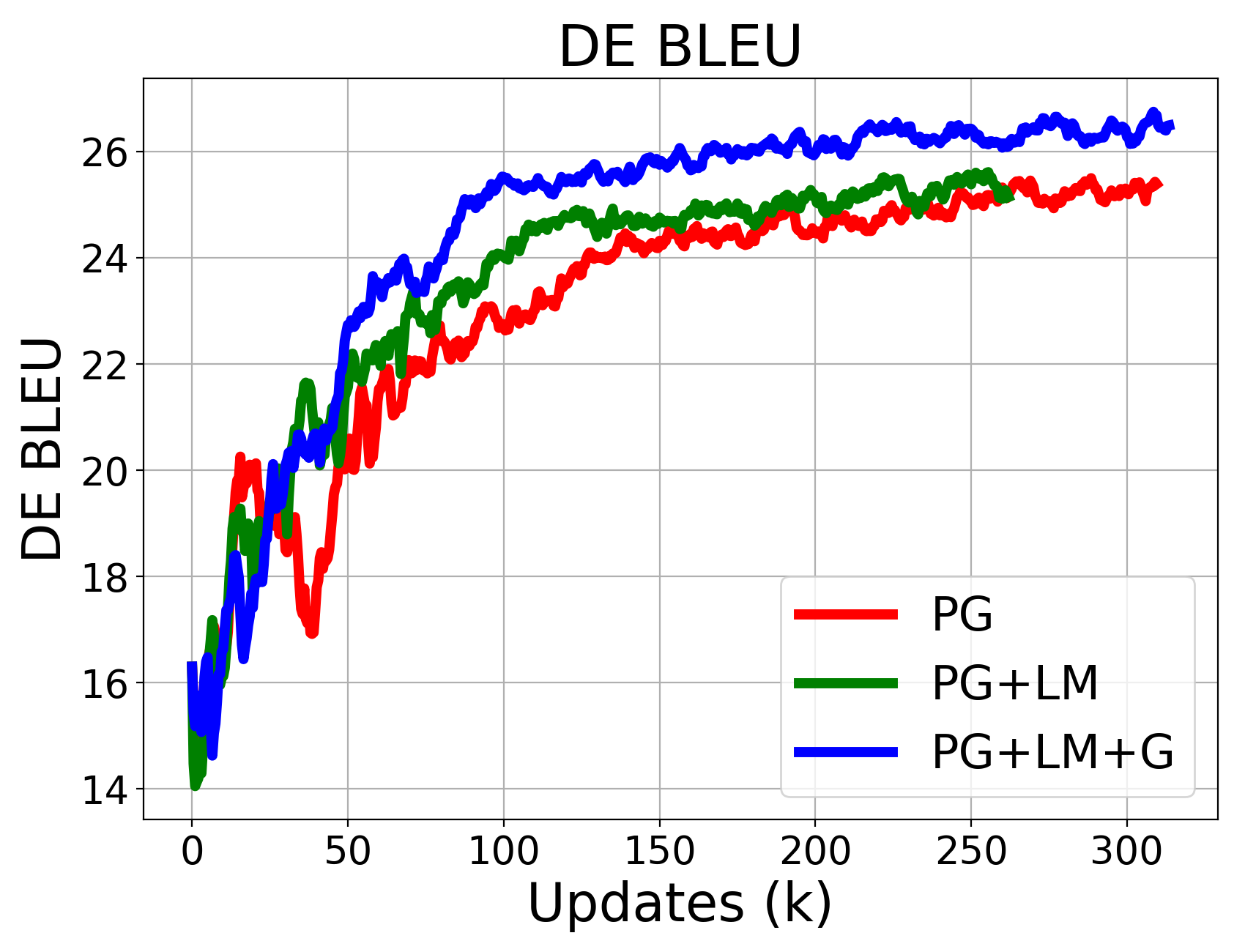}\hfill
\includegraphics[width=.28\textwidth]{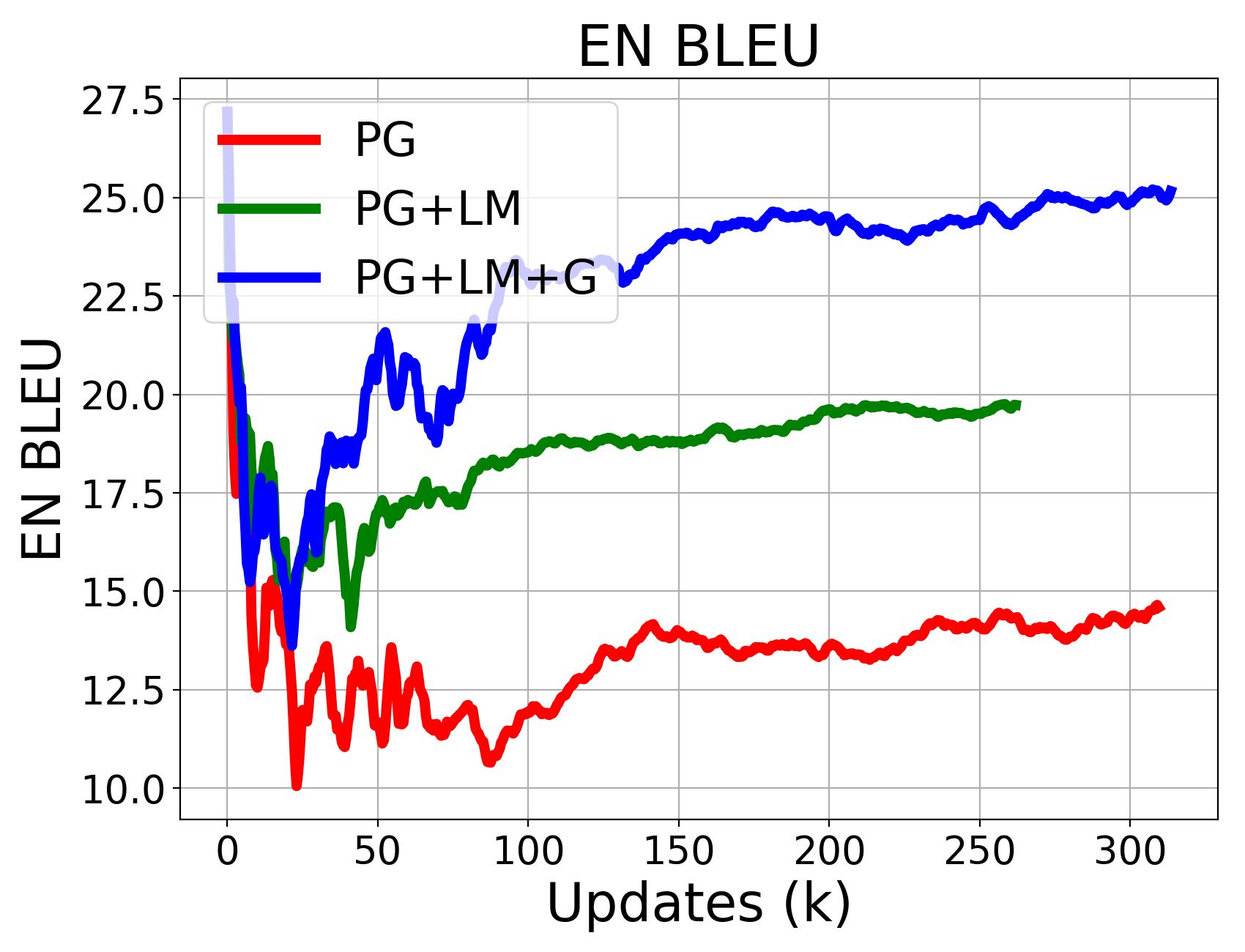}\hfill
\includegraphics[width=.28\textwidth]{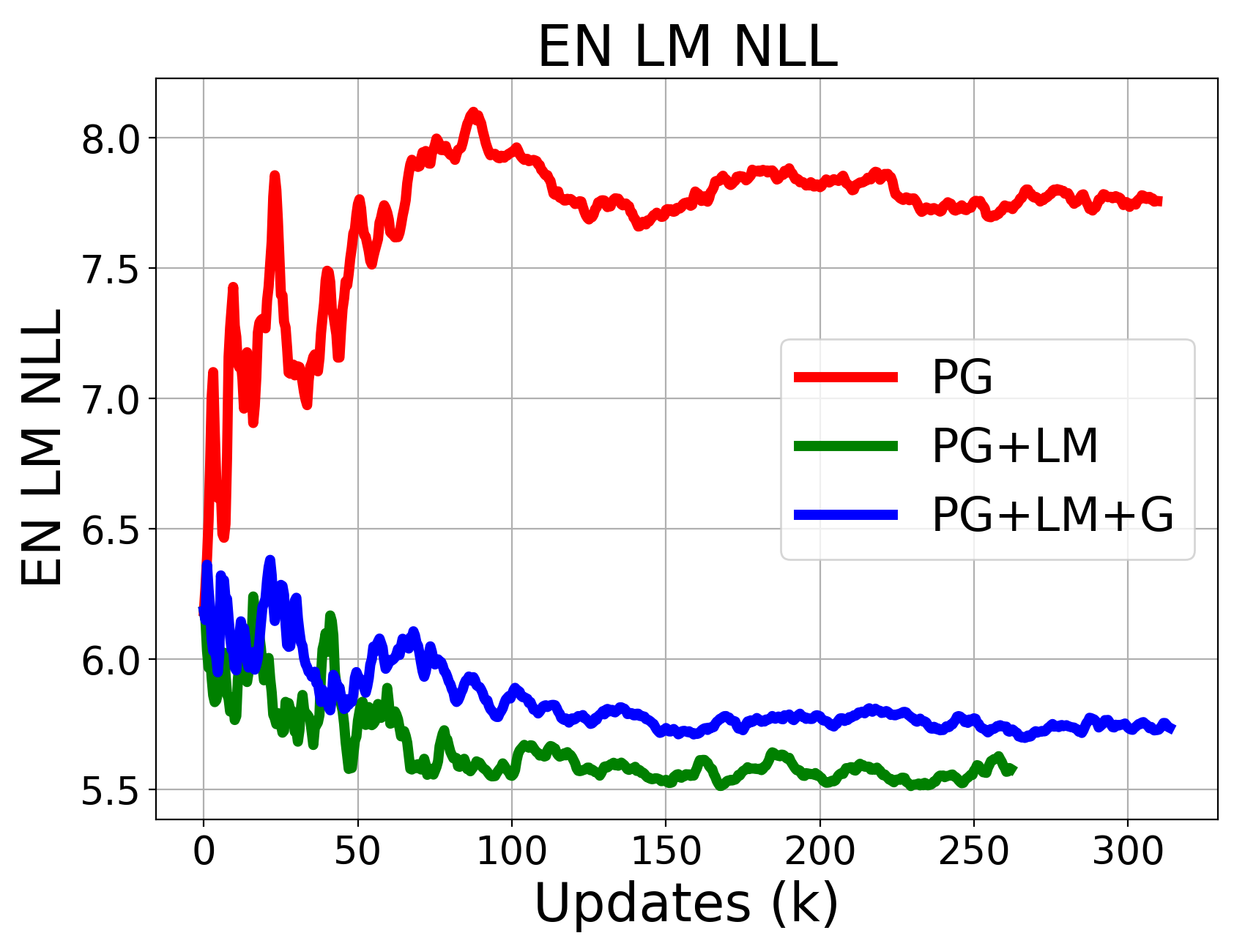}
\caption{Test set performance over time. En LM NLL curves show the NLL of English messages, computed by a language model trained on WikiText103. Lower En BLEU and higher En LM NLL indicate more language drift.}
\label{fig:wikiplot}
\end{figure*}

Meanwhile, we also train an NMT model of the same architecture and size directly on the Fr\rar De task in Multi30k Task 1 (without English intermediary). This serves as an upper bound on the Fr\rar De performance achievable with available data.

\section{Quantitative Results}

In Table~\ref{tab:quan_results}, the top three rows are the baselines described above. The pretrained-only baseline performs relatively poorly on Fr\rar De, conceivably because it was pretrained on a different corpus in a different domain (IWLST dataset is compiled from TED talks, while Multi30k dataset is a collection of image captions). Ensembling multiple English hypotheses for Agent B gives a negligible increase in Fr\rar De performance. When only Agent B is fine-tuned and Agent A is kept fixed, we observe an increase from 16.30 to 22.37 in Fr\rar De. Unsurprisingly, the upper bound NMT model directly trained end-to-end on Multi30k Fr\rar De (without any pivot, at the bottom of the table) performs best.

When the joint system is fine-tuned on German log-likelihood with policy gradients (PG), we observe a large, 8 BLEU increase in Fr\rar De (16.30\rar24.51) at the cost of a substantial, 15 BLEU drop in Fr\rar En (27.18\rar12.38). This clearly shows that optimizing for external reward may improve performance on that metric, but at the expense of a drastic language drift in the communication channel on which the reward is imposed.

When the system is fine-tuned only on staying grounded but without any language model constraint (PG+G), we obtain small performance improvements. 
This makes sense, since BLEU first and foremost focuses on the surface form.
When the agent is trained with the language model constraint (PG+LM), we notice a significant improvement in Fr\rar En BLEU. When the LM is trained on WikiText103, a widely used language modeling dataset, we observe an improvement of 9 BLEU score over PG (12.38\rar21.63). When the training corpus is closer to the target domain, such as MS COCO or Flickr30k, we observe even bigger increases. Fr\rar De translation also improves by 2--3 BLEU (24.51\rar26.88-27.67).

We see the biggest improvements in performance when agents are trained using both visual grounding feedback and the language model constraint (PG+LM+G). This is particularly pronounced with the LM trained on WikiText103: introducing visual grounding leads to more than 2 BLEU score improvement in Fr\rar En (21.63\rar23.65), and 1 BLEU score improvement in Fr\rar De (26.88\rar27.87). We hypothesize that the ``Englishness'' constraint forces agents to communicate with correct syntax and fluency, while the grounding model restricts the search space of languages to ones that are grounded in visual semantics. To investigate the contribution of grounding, we train a much stronger LM on all three datasets combined, and find that there is still more drift even with access to much more language modeling data (23.60\rar24.75).

\begin{table*}[t]
    \small
    \hfill
    \begin{minipage}{0.17\textwidth}
		\includegraphics[width=1.8cm]{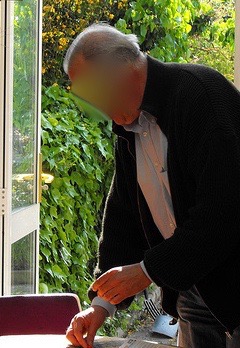}
		\vfill
		\vphantom{abc}
		\vphantom{abc}
		\includegraphics[width=1.8cm]{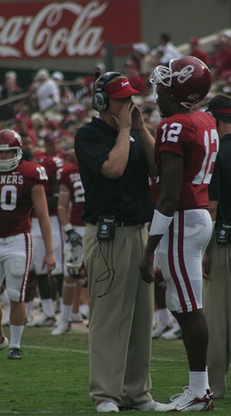}
    \end{minipage}
    \hspace{-1.2cm}
    \begin{minipage}{0.80\textwidth}
    \begin{center}
    \begin{tabular}{ l l l } \toprule
    
        \multirow{3}{*}{Ref} & Fr & {\color{red} un vieil homme} v\^etu d'une veste noire regarde sur la table \\
            & De & {\color{red} ein alter mann} in einer schwarzen jacke blickt auf den tisch \\
            & En & {\color{red} an old man} wearing a black jacket is looking on the table \\ \cmidrule{2-3}
            
        \multirow{3}{*}{En}  & PG & {\color{red} a old teaching} black watching on the table table table table table table\\ 
            & +LM & {\color{red} a old man} in a jacket looking on the table . " "  \\
            & +G & {\color{red} an old man} in a black jacket looking on the table .  \\ \cmidrule{2-3}
            
        \multirow{3}{*}{De}  & PG & {\color{red} ein \"alterer mann} in einem schwarzen hemd schaut auf den tisch . \\
            & +LM & {\color{red} ein alter mann} in einer jacke beobachtet einen tisch . \\
            & +G & {\color{red} ein \"alterer mann} in einer schwarzen jacke schaut auf den tisch .  \\ 
            
            \midrule
            
        \multirow{3}{*}{Ref} & Fr & un joueur de football am\'ericain en {\color{red} blanc et rouge} parle \`a un entra\^ineur . \\
            & De & ein {\color{red} rot-wei{\ss}} gekleideter footballspieler spricht mit einem trainer . \\
            & En & a football player in {\color{red} red and white} is talking to a coach . \\ \cmidrule{2-3}
            
        \multirow{3}{*}{En}  & PG & a player football american football american and {\color{red} red} talking talking a coach \\
            & +LM & a player of {\color{red} white and red} talking to a coach . " " " \\
            & +G & a football player in {\color{red} white and red} talking to a coach . \\ \cmidrule{2-3}
            
        \multirow{3}{*}{De}  & PG & ein footballspieler spricht mit einem spieler in einem {\color{red} roten} trikot . \\
            & +LM & ein {\color{red} wei{\ss}} gekleideter fu{\ss}ballspieler spricht zu einem trainer . \\
            & +G & ein fu{\ss}ballspieler in einem {\color{red} rot-wei{\ss}en} trikot spricht mit einem trainer .  \\
            
            \bottomrule
    \end{tabular}
    \end{center}
    \end{minipage}
    \captionof{table}{Two communication examples on the data from the Multi30k development set with different models (PG, PG+LM, PG+LM+G). The top three rows list the ground truth sentences, the middle three rows are the English messages sent by the Fr\rar En agent, and the bottom three rows show the German output from the En\rar De agent. We also show the corresponding images, which are only used to train the grounding model.}
    \label{tab:qual_results1}
    
\end{table*}

\begin{table}[t]
\small
\centering
\begin{center}
\begin{tabular}{ l p{5.5cm} } \toprule
    Reference & , . the and to of a that i in is it you we \&apos;s this {\color{red} \&quot;} \\%
    Pretrained & , the . to of and a i that in it we you \&apos;s is this {\color{red} \&quot;} was \\%
    PG & a the and , . in i {\color{red} \&quot;} this of to is we you ? that not for \\%
    PG+LM & the {\color{red} \&quot;} , of . and in a to this is i es you for we that with \\%
    PG+LM+G & the , . of a and to in is i this es we for that you at what \\

    \bottomrule
    \end{tabular}
    \caption{\label{tab:frequent_toks}Top 20 most frequent tokens (sorted) in English reference (Reference) or the output from Fr\rar En models.}
    \end{center}
\end{table}

\begin{table}[t]
    \small
    \centering
    \begin{tabular}{l p{0.4cm}p{0.4cm}p{0.4cm} p{0.4cm}p{0.4cm}p{0.4cm}p{0.4cm} } \toprule
            & \multicolumn{3}{c}{Function words} & \multicolumn{4}{c}{Content words} \\
            \cmidrule(lr){2-4} \cmidrule(lr){5-8}
            & TO & . & DT  & N & V & Adj & Adv \\ \midrule
    PG      & .22 & .36     & .57  & .38  & .17 & .32 & .26  \\
    PG+LM   & .55 & .84     & .72  & .39  & .18 & .21 & .25 \\
    PG+LM+G & .62 & .88     & .74  & .43  & .26 & .33 & .29 \\ 
    \bottomrule
    \end{tabular}
    \caption{Exact-match word recall by POS-tag on IWSLT development set: when the English reference contains a word of a certain POS tag, how often does the agent produce it.
    }
    \label{tab:pos}
\end{table}

\begin{table}[h]
    \small
    \centering
    \begin{tabular}{l p{0.6cm} p{0.5cm} p{0.5cm} p{0.6cm} p{0.5cm} p{0.5cm}} \toprule
            & \multicolumn{3}{c}{IWSLT} & \multicolumn{3}{c}{Multi30k} \\
            \cmidrule(lr){2-4} \cmidrule(lr){5-7}
            & unique    & /sent  & /all &  unique & /sent & /all \\ \midrule
    Reference     & 5,303  & 19.7  & 0.86   & 3,046  & 11.9 & 0.91\\ 
    Pretrained    & 4,657  & 17.9  & 0.85   & 2,867  & 12.0 & 0.87\\ \midrule
    PG      & 4,933  & 13.6  & 0.56 & 3,197  & 9.2  & 0.65  \\
    PG+LM   & 3,819  & 14.6  & 0.61 & 2,438  & 10.9 & 0.78 \\
    PG+LM+G & 4,327  & 15.7  & 0.74 & 2,550  & 10.7 & 0.84\\
    \bottomrule
    \end{tabular}
    \caption{Additional token frequency analysis. unique: the number of unique English tokens used in the whole development set. /sent: the number of unique English tokens used per sentence. /all: (the number of unique English tokens / the number of all English tokens.) }
    \label{tab:tokfreq}
\end{table}

It is important to check that the improvement from grounding is actually significant, so we perform a bootstrapped Wilcoxon signed-rank test~\citep{Wilcoxon:45} on paired English hypotheses for each reference sentence between PG+LM and PG+LM+G, using the model instance that gives the median communication performance (Fr\rar En\rar De BLEU) out of three runs. 
We assess significance on a bootstrapped test set (repeatedly sampled with replacement) and average the statistic over bootstrap samples. With the threshold of $p<0.02$, PG+LM+G is found to differ significantly for all the LM models, including the All model that had access to much more data. See Table \ref{tab:wilcoxon}.

\begin{figure}[t]
    \centering
    \includegraphics[width=.35\textwidth]{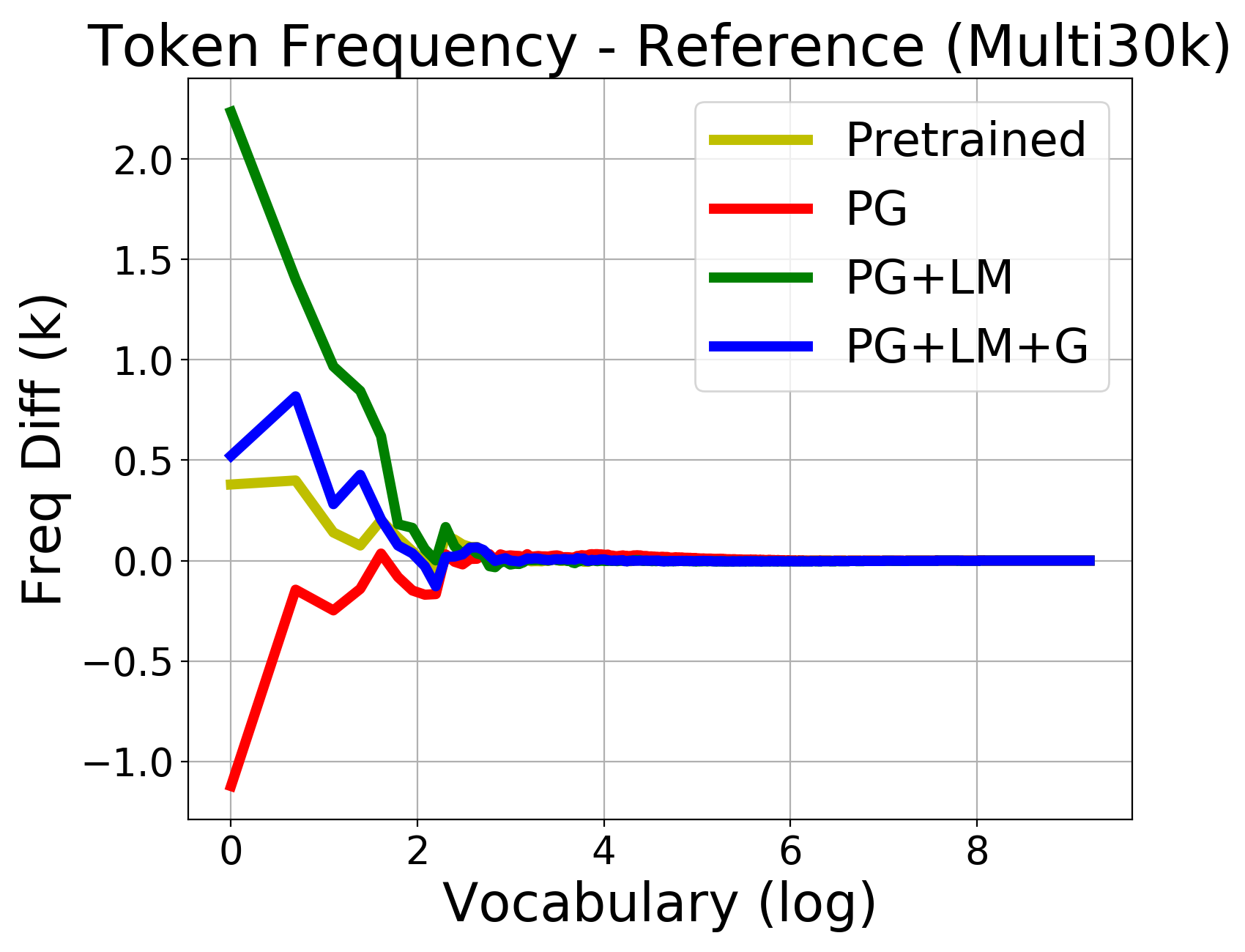}
    \caption{Token frequency analysis on three different models (PG, PG+LM, PG+LM+G) together with the pre-trained model before fine-tuning (Pretrained). We show word frequency curves for each model, after subtracting the reference English frequency statistics (both sorted in decreasing order). Positive y values indicate higher frequency values than the English reference, and negative y values indicate lower frequency values than English. The y-axis is the frequency difference in the thousands, and the x-axis shows the vocabulary index (sorted by frequency) in log scale.}
    \label{fig:tokfreq}
\end{figure}

Figure~\ref{fig:wikiplot} shows the learning curves, as measured by Fr\rar De BLEU (left), Fr\rar En BLEU (middle) and English LM negative log-likelihood (NLL; right). All models improve in fine-tuned task performance (left plot). We observe that vanilla PG fine-tuning quickly leads to highly ``un-English'' communication, as can be seen from a distinct increase in LM negative log-likelihood (right plot). While PG+LM achieves slightly lower LM NLL than PG+LM+G, its communication protocol drifts much more from English (middle plot). That is, for PG+LM, syntactic conformity is obtained at the expense of semantic preservation. Imposing both syntactic and semantic constraints makes models the least susceptible to drift, almost recovering to the original BLEU score (blue line, middle plot).

\section{Analysis}

\begin{table*}[t]
    \small
    \hfill
    \begin{minipage}{0.25\textwidth}
		\includegraphics[width=2.6cm]{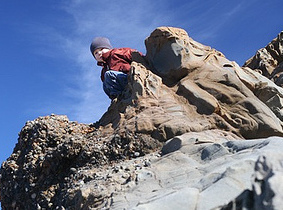}
		\includegraphics[width=2.6cm]{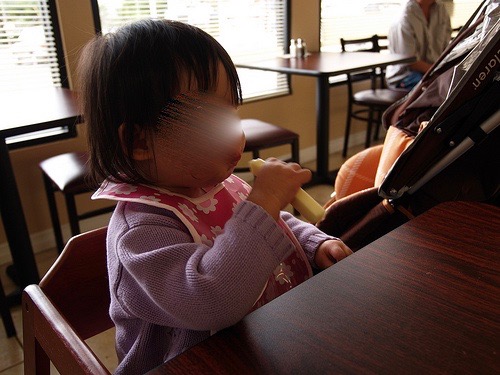}
    \end{minipage}
    \hspace{-2.3cm}
    \begin{minipage}{0.80\textwidth}
    \begin{center}
    \begin{tabular}{ l l } \toprule
        Fr src & un enfant assis sur un rocher. \\
        En ref & a {\color{red} child} sitting on a rock formation. \\
        En hyp & a {\color{red} punk} sitting sitting on on a broken\\
        De ref & ein kind sitzt auf einem felsen .\\
        De hyp & ein kind sitzt auf einem felsen .\\ \midrule
        
        Fr src & un petit enfant est assis \`{a} une table, en train de manger un go\^{u}ter. \\
        En ref & a {\color{red} toddler} is sitting at a table eating a snack . \\
        En hyp & a {\color{red} punk} sits sitting sitting next next a airline\\
        De ref & ein kleines kind sitzt an einem tisch und isst einen snack .\\
        De hyp & ein kind sitzt an einem tisch und liest ein buch . \\ 
    \bottomrule
    \end{tabular}
    \end{center}
    \end{minipage}
    \captionof{table}{Evidence of token flipping. The agents use the word ``punk'' to denote ``child'' or ``baby'', which is clearly not desirable.}
    \label{tab:qual_results2}
\end{table*}

A close investigation into the token statistics of each communication strategy reveals that PG fine-tuning causes the word frequency distribution to be flatter (see Figure~\ref{fig:tokfreq}). The PG model has negative frequency difference values for the most frequent tokens, indicating that PG downweighs frequent words severely, possibly because they are less discriminative. On the other hand, PG+LM gives highly positive frequency differences, meaning that language modeling alone disproportionately emphasizes frequent tokens. Using both the LM and grounding constraints keep the token frequencies closest to the pretrained regimes. Investigating the top 20 most frequent words shows that PG+LM disproportionately favors quotation marks, which are very common in many language modeling datasets but rare in Multi30k (see Table~\ref{tab:frequent_toks}).

Table~\ref{tab:pos} compares the degree of drift by part-of-speech, and shows that the PG model has very low recall on function words, such as periods and infinitives. Models trained with LM and grounding losses retain function words with much higher accuracy. PG fares relatively better with content words (nouns and verbs), but adding LM and grounding losses still outperform PG. Grounding leads to overall improvements in recall, particularly with content words. Conceivably, when optimizing Agent A's policy on the communication task alone, it is most crucial to relay content information to Agent B, and this might cause agents to ignore syntactic conformity in the original intermediate language. Imposing both syntactic and semantic constraints reduces the space of the intermediate communication protocol to a more stable language space, as reflected in overall task performance.

Table~\ref{tab:tokfreq} corroborates the finding that vanilla PG fine-tuning leads to flatter token frequency distributions, as the number of unique tokens used by PG is greater than that of the pretrained model. Despite using a more diverse set of tokens, PG uses the smallest number of unique symbols per sentence (/sent) and overall (/all). This implies that PG communication is redundant. PG+LM uses fewer tokens overall, and learns a sharper distribution using a smaller set of high-frequency tokens. Using both constraints yields a frequency distribution that most closely resembles the original one.

\section{Qualitative Results}

In the first example of Table~\ref{tab:qual_results1} (previous page), it is clear that PG's communication messages have significantly diverged from English: the model is highly repetitive (``table table table table table'') and misses some key content words such as ``man''. Agent B, however, correctly generates the German word ``mann''. This exemplifies a communication protocol that is successful in solving the task it was trained on, but not fully interpretable to humans. While the output from PG+LM is better, the grounded model's message (PG+LM+G) is distinctly the most fluent and semantically correct. 

In the second example, observe that the PG Agent B misinterprets ``talking talking a coach a coach'' as ``spricht mit einem spieler'' (talking to a player). The PG+LM+G model again generates a flawless English sentence. Furthermore, its agents succeed in communicating both colors (red and white) to German while retaining the original English words, when the other models fail to do so.

Interestingly, we observe some instances of token flipping with the PG model and to a lesser extent with the PG+LM model. For example, one particular model uses ``punk'' to describe ``child'' (see Table~\ref{tab:qual_results2}). As no occurrence of ``punk'' in any training data is associated with ``child'', the agents must have acquired this new meaning assignment during fine-tuning. Among 35 examples in the Multi30k development set where the English reference contains ``child'', the model uses ``punk'' 15 times, indicating this is no random phenomenon. We did not observe such examples with the PG+LM+G model.

\section{Conclusion}

In this paper, we show that language drift happens when fine-tuning natural language agents with some external reward using policy gradients without constraints. We investigate what constraints to put on the communication channel in order to mitigate this. We find that imposing syntactic constraints (via adding language model log-likelihood to the reward) does somewhat mitigate drift, but does not preserve semantic correspondence. We then observe that additionally imposing semantic constraints, e.g. with a perceptual grounding loss, yields communication protocols that best retain the original syntax and intended semantics, while giving the overall best communication performance.

Further analysis into the learned communication protocols reveals that pure PG fine-tuning tends to learn flatter and repetitive token distributions, while encouraging naturalness under a language model disproportionately emphasizes frequent syntactic tokens, yielding a much sharper token distribution than a natural language. The grounded model best retains the original token frequencies.



We examined language drift within a translation game as this allows for direct measurements at each step (input, intermediate, output), in a way where the semantics stays identical (i.e., the meaning is exactly the same for all languages and modalities) while the communication channel gets only an extrinsic reward (i.e., communication success). The findings in this work, however, are generally applicable to policy gradient fine-tuning of generative language models.
We believe that our work shows an intuitive method for addressing language drift and hope that it opens up interesting directions for future work.

\section*{Acknowledgments}

We thank the anonymous reviewers for their helpful feedback. We are grateful for support by eBay and NVIDIA. This work was partly supported by Samsung Advanced Institute of Technology (Next Generation Deep Learning: from pattern recognition to AI), Samsung Electronics (Improving Deep Learning using Latent Structure).

\bibliography{emnlp-ijcnlp-2019}

\begin{thebibliography}{51}
\expandafter\ifx\csname natexlab\endcsname\relax\def\natexlab#1{#1}\fi

\bibitem[{Andreas et~al.(2017)Andreas, Dragan, and Klein}]{Andreas:17}
Jacob Andreas, Anca~D. Dragan, and Dan Klein. 2017.
\newblock Translating neuralese.
\newblock In \emph{Proceedings of the 55th Annual Meeting of the Association
  for Computational Linguistics}, pages 232--242.

\bibitem[{Bahdanau et~al.(2016)Bahdanau, Brakel, Xu, Goyal, Lowe, Pineau,
  Courville, and Bengio}]{Bahdanau:16}
Dzmitry Bahdanau, Philemon Brakel, Kelvin Xu, Anirudh Goyal, Ryan Lowe, Joelle
  Pineau, Aaron~C. Courville, and Yoshua Bengio. 2016.
\newblock An actor-critic algorithm for sequence prediction.
\newblock \emph{arXiv preprint arXiv:1607.07086}.

\bibitem[{Bahdanau et~al.(2015)Bahdanau, Cho, and Bengio}]{Bahdanau:15}
Dzmitry Bahdanau, Kyunghyun Cho, and Yoshua Bengio. 2015.
\newblock Neural machine translation by jointly learning to align and
  translate.
\newblock In \emph{Proceedings of the International Conference on Learning
  Representations (ICLR)}.

\bibitem[{Baroni(2016)}]{Baroni:16}
Marco Baroni. 2016.
\newblock Grounding distributional semantics in the visual world.
\newblock \emph{Language and Linguistics Compass}, 10(1):3--13.

\bibitem[{Briscoe(2002)}]{Briscoe:02}
Ted Briscoe. 2002.
\newblock \emph{Linguistic evolution through language acquisition}.
\newblock Cambridge University Press.

\bibitem[{Cettolo et~al.(2012)Cettolo, Girardi, and
  Federico}]{Cettolo:2012iwslt}
Mauro Cettolo, Christian Girardi, and Marcello Federico. 2012.
\newblock Wit$^3$: Web inventory of transcribed and translated talks.
\newblock In \emph{Proceedings of the 16$^{th}$ Conference of the European
  Association for Machine Translation (EAMT)}, pages 261--268, Trento, Italy.

\bibitem[{Chen et~al.(2018)Chen, Liu, and Li}]{Chen:18}
Yun Chen, Yang Liu, and Victor O.~K. Li. 2018.
\newblock Zero-resource neural machine translation with multi-agent
  communication game.
\newblock In \emph{Proceedings of the Thirty-Second {AAAI} Conference on
  Artificial Intelligence}.

\bibitem[{Cheng et~al.(2017)Cheng, Yang, Liu, Sun, and Xu}]{Cheng:17}
Yong Cheng, Qian Yang, Yang Liu, Maosong Sun, and Wei Xu. 2017.
\newblock Joint training for pivot-based neural machine translation.
\newblock In \emph{Proceedings of the Twenty-Sixth International Joint
  Conference on Artificial Intelligence}, pages 3974--3980.

\bibitem[{Cho et~al.(2014)Cho, Van~Merri{\"e}nboer, Gulcehre, Bahdanau,
  Bougares, Schwenk, and Bengio}]{Cho:14gru}
Kyunghyun Cho, Bart Van~Merri{\"e}nboer, Caglar Gulcehre, Dzmitry Bahdanau,
  Fethi Bougares, Holger Schwenk, and Yoshua Bengio. 2014.
\newblock Learning phrase representations using rnn encoder-decoder for
  statistical machine translation.
\newblock \emph{arXiv preprint arXiv:1406.1078}.

\bibitem[{Chrupa{\l}a et~al.(2015)Chrupa{\l}a, K{\'a}d{\'a}r, and
  Alishahi}]{Chrupala:15}
Grzegorz Chrupa{\l}a, Akos K{\'a}d{\'a}r, and Afra Alishahi. 2015.
\newblock Learning language through pictures.
\newblock \emph{arXiv preprint arXiv:1506.03694}.

\bibitem[{{Elliott} et~al.(2016){Elliott}, {Frank}, {Sima'an}, and
  {Specia}}]{Elliott:16}
Desmond {Elliott}, Stella {Frank}, Khalil {Sima'an}, and Lucia {Specia}. 2016.
\newblock Multi30k: Multilingual english-german image descriptions.
\newblock In \emph{Proceedings of the 5th Workshop on Vision and Language},
  pages 70--74.

\bibitem[{Elliott and K{\'a}d{\'a}r(2017)}]{Elliott:17}
Desmond Elliott and Akos K{\'a}d{\'a}r. 2017.
\newblock Imagination improves multimodal translation.
\newblock \emph{arXiv preprint arXiv:1705.04350}.

\bibitem[{Evtimova et~al.(2018)Evtimova, Drozdov, Kiela, and Cho}]{Evtimova:18}
Katrina Evtimova, Andrew Drozdov, Douwe Kiela, and Kyunghyun Cho. 2018.
\newblock Emergent language in a multi-modal, multi-step referential game.
\newblock In \emph{Proceedings of the International Conference on Learning
  Representations}.

\bibitem[{Faghri et~al.(2018)Faghri, Fleet, Kiros, and Fidler}]{Faghri:18}
Fartash Faghri, David~J. Fleet, Jamie Kiros, and Sanja Fidler. 2018.
\newblock {VSE++:} improving visual-semantic embeddings with hard negatives.
\newblock In \emph{British Machine Vision Conference}, page~12.

\bibitem[{Firat et~al.(2016)Firat, Sankaran, Al-Onaizan, Vural, and
  Cho}]{firat2016zero}
Orhan Firat, Baskaran Sankaran, Yaser Al-Onaizan, Fatos T~Yarman Vural, and
  Kyunghyun Cho. 2016.
\newblock Zero-resource translation with multi-lingual neural machine
  translation.
\newblock \emph{arXiv preprint arXiv:1606.04164}.

\bibitem[{Foerster et~al.(2016)Foerster, Assael, de~Freitas, and
  Whiteson}]{foerster2016learning}
Jakob Foerster, Ioannis~Alexandros Assael, Nando de~Freitas, and Shimon
  Whiteson. 2016.
\newblock Learning to communicate with deep multi-agent reinforcement learning.
\newblock In \emph{Advances in Neural Information Processing Systems}, pages
  2137--2145.

\bibitem[{Gauthier and Mordatch(2016)}]{gauthier16paradigm}
Jon Gauthier and Igor Mordatch. 2016.
\newblock A paradigm for situated and goal-driven language learning.
\newblock \emph{arXiv preprint arXiv:1610.03585}.

\bibitem[{Gella et~al.(2017)Gella, Sennrich, Keller, and Lapata}]{Gella:17}
Spandana Gella, Rico Sennrich, Frank Keller, and Mirella Lapata. 2017.
\newblock Image pivoting for learning multilingual multimodal representations.
\newblock \emph{arXiv preprint arXiv:1707.07601}.

\bibitem[{Gu et~al.(2017)Gu, Cho, and Li}]{Gu:17}
Jiatao Gu, Kyunghyun Cho, and Victor O.~K. Li. 2017.
\newblock Trainable greedy decoding for neural machine translation.
\newblock In \emph{Proceedings of the 2017 Conference on Empirical Methods in
  Natural Language Processing}, pages 1968--1978.

\bibitem[{Havrylov and Titov(2017)}]{Havrylov:17}
Serhii Havrylov and Ivan Titov. 2017.
\newblock Emergence of language with multi-agent games: Learning to communicate
  with sequences of symbols.
\newblock In \emph{Advances in Neural Information Processing Systems 30: Annual
  Conference on Neural Information Processing Systems}, pages 2146--2156.

\bibitem[{He et~al.(2016)He, Zhang, Ren, and Sun}]{He:16}
Kaiming He, Xiangyu Zhang, Shaoqing Ren, and Jian Sun. 2016.
\newblock Deep residual learning for image recognition.
\newblock In \emph{2016 {IEEE} Conference on Computer Vision and Pattern
  Recognition, {CVPR}}, pages 770--778.

\bibitem[{Hitschler et~al.(2016)Hitschler, Schamoni, and
  Riezler}]{Hitschler:16}
Julian Hitschler, Shigehiko Schamoni, and Stefan Riezler. 2016.
\newblock Multimodal pivots for image caption translation.
\newblock \emph{arXiv preprint arXiv:1601.03916}.

\bibitem[{K{\'a}d{\'a}r et~al.(2018)K{\'a}d{\'a}r, Elliott, C{\^o}t{\'e},
  Chrupa{\l}a, and Alishahi}]{Kadar:18}
{\'A}kos K{\'a}d{\'a}r, Desmond Elliott, Marc-Alexandre C{\^o}t{\'e}, Grzegorz
  Chrupa{\l}a, and Afra Alishahi. 2018.
\newblock Lessons learned in multilingual grounded language learning.
\newblock \emph{arXiv preprint arXiv:1809.07615}.

\bibitem[{Kiela(2017)}]{Kiela:17thesis}
Douwe Kiela. 2017.
\newblock \emph{{Deep Embodiment: Grounding Semantics in Perceptual
  Modalities}}.
\newblock Ph.D. thesis, University of Cambridge, Computer Laboratory.

\bibitem[{Kiela et~al.(2017)Kiela, Conneau, Jabri, and
  Nickel}]{Kiela:17groundsent}
Douwe Kiela, Alexis Conneau, Allan Jabri, and Maximilian Nickel. 2017.
\newblock Learning visually grounded sentence representations.
\newblock \emph{arXiv preprint arXiv:1707.06320}.

\bibitem[{Kingma and Ba(2014)}]{Kingma:14}
Diederik~P. Kingma and Jimmy Ba. 2014.
\newblock Adam: {A} method for stochastic optimization.
\newblock \emph{CoRR}.

\bibitem[{Kirby(2001)}]{Kirby:01}
Simon Kirby. 2001.
\newblock Spontaneous evolution of linguistic structure-an iterated learning
  model of the emergence of regularity and irregularity.
\newblock \emph{IEEE Transactions on Evolutionary Computation}, 5(2):102--110.

\bibitem[{Koehn et~al.(2007)Koehn, Hoang, Birch, Callison{-}Burch, Federico,
  Bertoldi, Cowan, Shen, Moran, Zens, Dyer, Bojar, Constantin, and
  Herbst}]{Koehn:07}
Philipp Koehn, Hieu Hoang, Alexandra Birch, Chris Callison{-}Burch, Marcello
  Federico, Nicola Bertoldi, Brooke Cowan, Wade Shen, Christine Moran, Richard
  Zens, Chris Dyer, Ondrej Bojar, Alexandra Constantin, and Evan Herbst. 2007.
\newblock Moses: Open source toolkit for statistical machine translation.
\newblock In \emph{Proceedings of the 45th Annual Meeting of the Association
  for Computational Linguistics}.

\bibitem[{Kottur et~al.(2017)Kottur, Moura, Lee, and Batra}]{Kottur:17}
Satwik Kottur, Jos{\'{e}} M.~F. Moura, Stefan Lee, and Dhruv Batra. 2017.
\newblock Natural language does not emerge 'naturally' in multi-agent dialog.
\newblock In \emph{Proceedings of the 2017 Conference on Empirical Methods in
  Natural Language Processing}, pages 2962--2967.

\bibitem[{Lazaridou et~al.(2016)Lazaridou, Peysakhovich, and
  Baroni}]{lazaridou2016multi}
Angeliki Lazaridou, Alexander Peysakhovich, and Marco Baroni. 2016.
\newblock Multi-agent cooperation and the emergence of (natural) language.
\newblock \emph{arXiv preprint arXiv:1612.07182}.

\bibitem[{Lazaridou et~al.(2017)Lazaridou, Peysakhovich, and
  Baroni}]{Lazaridou:17}
Angeliki Lazaridou, Alexander Peysakhovich, and Marco Baroni. 2017.
\newblock Multi-agent cooperation and the emergence of (natural) language.
\newblock In \emph{Proceedings of the International Conference on Learning
  Representations}.

\bibitem[{Lee et~al.(2018)Lee, Cho, Weston, and Kiela}]{Lee:18}
Jason Lee, Kyunghyun Cho, Jason Weston, and Douwe Kiela. 2018.
\newblock Emergent translation in multi-agent communication.
\newblock In \emph{Proceedings of the International Conference on Learning
  Representations}.

\bibitem[{Lewis et~al.(2017)Lewis, Yarats, Dauphin, Parikh, and
  Batra}]{Lewis:17}
Mike Lewis, Denis Yarats, Yann~N Dauphin, Devi Parikh, and Dhruv Batra. 2017.
\newblock Deal or no deal? end-to-end learning for negotiation dialogues.
\newblock \emph{arXiv preprint arXiv:1706.05125}.

\bibitem[{Li et~al.(2017)Li, Monroe, Shi, Jean, Ritter, and Jurafsky}]{Li:17}
Jiwei Li, Will Monroe, Tianlin Shi, S{\'{e}}bastien Jean, Alan Ritter, and Dan
  Jurafsky. 2017.
\newblock Adversarial learning for neural dialogue generation.
\newblock In \emph{Proceedings of the 2017 Conference on Empirical Methods in
  Natural Language Processing}, pages 2157--2169.

\bibitem[{Mordatch and Abbeel(2017)}]{Mordatch:17}
Igor Mordatch and Pieter Abbeel. 2017.
\newblock Emergence of grounded compositional language in multi-agent
  populations.
\newblock \emph{arXiv preprint arXiv:1705.10369}.

\bibitem[{Muraki(1986)}]{Muraki:86}
Kazunori Muraki. 1986.
\newblock {VENUS:} two-phase machine translation system.
\newblock \emph{Future Generation Comp. Syst.}, 2(2):117--119.

\bibitem[{Nakayama and Nishida(2017)}]{Nakayama:17}
Hideki Nakayama and Noriki Nishida. 2017.
\newblock Zero-resource machine translation by multimodal encoder--decoder
  network with multimedia pivot.
\newblock \emph{Machine Translation}, 31(1-2):49--64.

\bibitem[{Nogueira and Cho(2017)}]{Nogueira:17}
Rodrigo Nogueira and Kyunghyun Cho. 2017.
\newblock Task-oriented query reformulation with reinforcement learning.
\newblock In \emph{Proceedings of the 2017 Conference on Empirical Methods in
  Natural Language Processing}, pages 574--583.

\bibitem[{Nowak and Krakauer(1999)}]{Nowak:99}
Martin~A Nowak and David~C Krakauer. 1999.
\newblock The evolution of language.
\newblock \emph{Proceedings of the National Academy of Sciences},
  96(14):8028--8033.

\bibitem[{Paulus et~al.(2017)Paulus, Xiong, and Socher}]{Paulus:17}
Romain Paulus, Caiming Xiong, and Richard Socher. 2017.
\newblock A deep reinforced model for abstractive summarization.
\newblock \emph{arXiv preprint arXiv:1705.04304}.

\bibitem[{Radford et~al.(2019)Radford, Wu, Child, Luan, Amodei, and
  Sutskever}]{Radford:2019gpt}
Alec Radford, Jeffrey Wu, Rewon Child, David Luan, Dario Amodei, and Ilya
  Sutskever. 2019.
\newblock Language models are unsupervised multitask learners.

\bibitem[{Ranzato et~al.(2015)Ranzato, Chopra, Auli, and Zaremba}]{Ranzato:15}
Marc'Aurelio Ranzato, Sumit Chopra, Michael Auli, and Wojciech Zaremba. 2015.
\newblock Sequence level training with recurrent neural networks.
\newblock \emph{arXiv preprint arXiv:1511.06732}.

\bibitem[{Sennrich et~al.(2016)Sennrich, Haddow, and Birch}]{Sennrich:16}
Rico Sennrich, Barry Haddow, and Alexandra Birch. 2016.
\newblock Neural machine translation of rare words with subword units.
\newblock In \emph{Proceedings of the 54th Annual Meeting of the Association
  for Computational Linguistics}.

\bibitem[{Skyrms(2010)}]{Skyrms:10}
Brian Skyrms. 2010.
\newblock \emph{Signals: Evolution, learning, and information}.
\newblock Oxford University Press.

\bibitem[{Srivastava et~al.(2014)Srivastava, Hinton, Krizhevsky, Sutskever, and
  Salakhutdinov}]{Srivastava:14}
Nitish Srivastava, Geoffrey~E. Hinton, Alex Krizhevsky, Ilya Sutskever, and
  Ruslan Salakhutdinov. 2014.
\newblock Dropout: a simple way to prevent neural networks from overfitting.
\newblock \emph{Journal of Machine Learning Research}, 15(1):1929--1958.

\bibitem[{Steels(1997)}]{Steels:97}
Luc Steels. 1997.
\newblock The synthetic modeling of language origins.
\newblock \emph{Evolution of communication}, 1(1):1--34.

\bibitem[{Wang et~al.(2017)Wang, Zhao, Zhang, Zong, and Xue}]{Wang:17}
Yining Wang, Yang Zhao, Jiajun Zhang, Chengqing Zong, and Zhengshan Xue. 2017.
\newblock Towards neural machine translation with partially aligned corpora.
\newblock In \emph{Proceedings of the Eighth International Joint Conference on
  Natural Language Processing}, pages 384--393.

\bibitem[{Wilcoxon(1945)}]{Wilcoxon:45}
Frank Wilcoxon. 1945.
\newblock Individual comparisons by ranking methods.
\newblock \emph{Biometrics bulletin}, 1(6):80--83.

\bibitem[{Williams(1992)}]{Williams:92}
Ronald~J Williams. 1992.
\newblock Simple statistical gradient-following algorithms for connectionist
  reinforcement learning.
\newblock \emph{Machine learning}, 8(3-4):229--256.

\bibitem[{Young et~al.(2014)Young, Lai, Hodosh, and
  Hockenmaier}]{Young:2014flickr30k}
Peter Young, Alice Lai, Micah Hodosh, and Julia Hockenmaier. 2014.
\newblock From image descriptions to visual denotations: New similarity metrics
  for semantic inference over event descriptions.
\newblock \emph{Transactions of the Association for Computational Linguistics},
  2:67--78.

\bibitem[{Zoph and Knight(2016)}]{zoph2016multi}
Barret Zoph and Kevin Knight. 2016.
\newblock Multi-source neural translation.
\newblock \emph{arXiv preprint arXiv:1601.00710}.

\end{thebibliography}
\bibliographystyle{acl_natbib}

\end{document}